\documentclass[11pt]{article}
\usepackage{coling2020}
\usepackage{times}
\usepackage{url}
\usepackage{latexsym}
\usepackage{graphicx}
\usepackage{stfloats}
\usepackage{multirow}

\colingfinalcopy 

\title{Contextual Argument Component Classification for Class Discussions}

\author{Luca Lugini \and  {\bf Diane Litman}\\
         \\ Department of Computer Science and \\Learning Research and Development Center\\University of Pittsburgh \\ Pittsburgh, PA, USA}

\date{}

\begin{document}
\maketitle
\begin{abstract}
Argument mining systems often consider contextual information, i.e. information outside of an argumentative discourse unit, when trained to accomplish  tasks such as argument component identification, classification, and relation extraction.
However, prior work has not carefully analyzed  the utility of different contextual properties in context-aware models.
In this work, we show how two different types of contextual information, local discourse context and speaker context, can be incorporated into a computational model for classifying argument components  in 
multi-party classroom discussions. We find that both context types can improve performance, although the improvements are dependent on context size and position.
\end{abstract}

\section{Motivation}
\label{sec:introduction}
\blfootnote{
    \hspace{-0.65cm}  
    This work is licensed under a Creative Commons 
    Attribution 4.0 International License.
    License details:
    \url{http://creativecommons.org/licenses/by/4.0/}.
}
In a typical argument mining system, the first task is identifying spans of text consisting of argumentative discourse units (ADU), i.e. argument component identification (ACI).
The next task, argument component classification (ACC), consists of assigning a label to each ADU according to an argument model, e.g. claims, evidence, etc.
\begin{table*}[!b]
\vspace{-3mm}
\small
\centering
\begin{tabular}{c|c|p{12.2cm}|c}
\textbf{Row} & \textbf{ID} & \textbf{ADU} & \textbf{AC} \\ \hline
1 & 7 & I feel that she included herself in the book to give more of an unbiased point of view to it. & claim \\ \hline
2 & 7 & Because, obviously the family was going to be completely against the doctors and what they did, and all that. & evidence \\ \hline
3 & 7 & But I feel by her including herself in it, she did show both sides of the story. & warrant \\ \hline
4 & 10 & I'd argue against that keeping her as unbiased. Because I think it puts more of a bias on the Lacks family's side. & claim \\ \hline
5 & 7 & I also feel she included herself because this was such a big part of her career from learning all this. And she became attached to the family members. & claim \\ \hline
\end{tabular}
\caption{Excerpt from a classroom discussion showing speaker ID, ADU and argument component labels.}
\label{tab:example}
\vspace{-2mm}
\end{table*}
For example, row 1 in Table \ref{tab:example} is labeled claim since speaker 7 provides their personal view, while row 2 is labeled evidence because it references facts from a text.

While ``context'' has been used in the argument mining literature to refer to several phenomena, 
we consider context to be auxiliary textual information outside the span of an ADU.
It is generally accepted that context is important in argument mining.
Stab and Gurevych \shortcite{Stab:14b} as well as Nguyen and Litman \shortcite{Nguyen:16} 
use context features extracted from the sentence containing an ADU to improve ACC.
Persing and Ng \shortcite{Persing:16}, Habernal and Gurevych \shortcite{Habernal:17} and Aker et al. \shortcite{Aker:17}
similarly use contextual features in joint ACI/ACC models.
Optiz and Frank \shortcite{Optiz:19} analyze a previous argument mining system and find that, for its predictions, it relies on context more than it does on ADU content.
Chakrabarty et al. \shortcite{Chakrabarty:19} 
indirectly model context in ACC 
by fine-tuning a BERT model to predict the next sentence (i.e. the context). 
Eger et al. \shortcite{Eger:17} analyzed several neural models for jointly performing 
ACI, ACC and argument relation extraction.
All of these works share several limitations: \textit{(i)} the context is either limited to a single configuration (e.g. one sentence before/after the ADU) or optimized along a single dimension (typically size but not position); \textit{(ii)} only  a subset of 
the features extracted for the target ADU are also extracted for context; \textit{(iii)} 
context is typically based on ADU adjacency, although other ways of building  context (e.g. based on speakers in multi-party dialogues) are possible. 

In this work, we improve upon baseline ACC models for multi-party discussions by incorporating
two types of contexts. 
We define {\it local context} as ADUs preceding and/or following a target ADU, regardless of speaker.
{\it Speaker context} consists of previous ADUs that a specific speaker previously voiced during the discussion.
Our results show that both context types can individually and jointly improve ACC performance,  
with performance gains 
dependent on context size 
and position.

\section{Dataset and  Models}
\label{sec:dataset}
The dataset\footnote{Obtained from  ~\url{https://discussiontracker.cs.pitt.edu}} used to build and evaluate our proposed ACC models consists of 3,135 ADUs in a corpus of 29 text-based (i.e. centered around a book, play, or other literature piece), multi-party classroom discussions between high school 
students (average 15 students per discussion) \cite{olshefski:20}.
The discussions 
(average length of 34 minutes)
were audio-recorded, manually transcribed, and student turns were manually segmented into multiple ADUs when needed.
ADUs were then manually annotated according to a simplified version of Toulmin's argumentative model \cite{Toulmin:58} consisting of three labels: \textit{(i) claims}, arguable statements that voice a specific interpretation of a text; \textit{(ii) evidence}, facts or documentation used to support a claim; \textit{(iii) warrant}, reasoning given to explain how certain evidence supports a claim.
An inter-rater reliability analysis showed a Krippendorff $\alpha_U$ of 0.96 for segmentation and Cohen Kappa of 0.74 for argument components.
The dataset contains 3,135 ADUs: 65.3\% claims, 24.3\% evidence, and 10.4\% warrants.
ADUs are  additionally labeled with the ID of the speaker who voiced the utterance.
Table \ref{tab:example} shows an excerpt from an annotated discussion.

To evaluate the utility of context, we introduce two baseline ACC models and propose several contextual extensions.
The source code for all models (and all parameters) is available at ~\url{https://github.com/lucalugini/coling2020_argmining}.

\textbf{Baseline Models.} \quad
Our first model ({\it hybrid baseline}) is based on the  model of Lugini and Litman \shortcite{Lugini:18b} which was developed for a similar type of dataset,
where an embedding generated through a convolutional neural network (CNN) is concatenated to a set of handcrafted features, and a softmax classifier is used to predict argument components. 
Given the limited size of our dataset, however, we only use
a subset of the original model's handcrafted features, namely those used by Speciteller \cite{Li:15}; this reduces the number of handcrafted features from over 7,000 to 114 and avoids overfit.
The Speciteller feature set consists of pretrained word vectors (average of the word vectors for each word in the ADU), as well as number of connectives, number of words, number of numbers, number of symbols, number of capital letters, number of stopwords normalized by ADU length, number of subjective and polar words (from the MPQA \cite{Wilson:09} and the General Inquirer \cite{Stone:63} dictionaries), average word familiarity (from MRC Psycholinguistic Database \cite{Wilson:88}), average characters per word, and inverse document frequency statistics (minimum and maximum).
The dimensionality of the final feature vector is 2,514 (114 for the handcrafted features and 2,400 for the CNN).
Our second model ({\it BERT baseline}) is based on recent advances related to Transformer architectures \cite{vaswani:17}: a BERT pretrained model \cite{devlin:18,HuggingFacesTransformers} generates word embeddings of dimensionality 768; average pooling is used to compute a fixed-size ADU embedding; a softmax classifier predicts argument components.

{\bf Adding Local Context.} \quad
We define local context as ADUs preceding and/or following the target ADU, regardless of speaker ID.
Context {\it size} is measured in terms of complete ADUs (i.e. entire utterance or part of it), while context {\it position} refers to the relative position of the context ADUs to the target ADU (i.e. preceding, following, both).
We believe defining context in terms of ADUs is the most straightforward choice since it is the same unit of analysis used for individual argument components.
Though beyond the scope of this paper, another compelling choice consists in defining context based on the number of words outside the target ADU instead.
We address the prior work limitations highlighted earlier in two ways:
\textit{(i)} we explore the impact of varying both the size and position of ADUs included in the context, instead of picking a single position and optimizing size;
\textit{(ii)} we model context using the same features used for the target ADU.
Each context ADU is converted into a fixed-size feature vector using the baseline models described above, then concatenated to the feature vector for the target ADU.
A maximum context size of 6 was chosen based on results showing diminishing returns and on the fact that increasing size further would go beyond ``local'' context.
We additionally evaluate whether context size and position can be automatically optimized by adding an attention layer \cite{luong:15}: context size is set to the maximum value and
both preceding and following positions are included; the attention mechanism then aggregates all context ADUs into a single  vector.

\textbf{Adding Speaker Context.} \quad
Students exhibit highly variable behavior with respect to how they build arguments. 
For example, in a discussion between six students from the dataset, one student only voiced claims, only two  students voiced warrants, and only two students voiced more than 10\% of their argument components as evidence.
We hypothesize that the argument component classifier can benefit from being informed of the propensity of a particular speaker to voice each argument component at any given point. 
While we have access to speaker ID,  when making predictions ground truth ADU labels are not available; therefore we need to extract information from the ADU text.
Given the speaker ID for the target ADU, the speaker context module performs the following steps: (1) gather the speaker's previous ADUs from the discussion; (2) convert each ADU into a feature vector; (3) aggregate them into a single, fixed-size feature vector and concatenate it with the baseline (and possibly with the local context).
Step 1 involves simply filtering out ADUs based on speaker ID, which is readily available in each discussion.
Step 2 can be achieved in several ways, however, for the sake of simplicity, in the hybrid baseline we decided to use a CNN to generate a feature vector for each ADU.
In order to further reduce complexity 
we implemented a CNN with the same structure as the one in the hybrid baseline model, but with the number of filters reduced from 16 to 4. This resulted in a 200-dimensional vector.
For the BERT baseline the same embedding - average pooling model was used in this step.
Step 3 was accomplished using a Long Short-Term Memory (LSTM) network \cite{LSTM:97}.
The final speaker context feature vector has dimensionality 100 (output dimensionality of the LSTM). 
We additionally experimented with automatically optimizing speaker context size by replacing the LSTM with an attention layer: setting the speaker context size to the maximum (40 in this case) the attention layer aggregates all ADUs into a single feature vector.

\textbf{Context Examples.} \quad
Assume the target ADU is row 5 in Table \ref{tab:example}. A local context of size  
3 contains rows 2, 3, and 4.  A speaker context of size 
3 contains rows 1, 2 and 3.

\section{Experiments and Results}
\label{sec:results}
All models are evaluated using ten fold cross-validation, and results are shown in Table \ref{tab:results}.
\begin{table*}[!b]
\small
\centering
\begin{tabular}{c|c|c|c|c|c|c}
\textbf{Row} & \textbf{Model} & \textbf{Context} & \textbf{Kappa} & \textbf{Precision} & \textbf{Recall} & \textbf{F-score} \\ \hline
1 & \multirow{4}{*}{Hybrid Baseline} & - & 0.350 & 0.535 & 0.531 & 0.509\\ \cline{1-1} \cline{3-7}
2 &  & Local Context & 0.521 & 0.657 & \textbf{0.727} & 0.676\\ \cline{1-1} \cline{3-7}
3 &  & Speaker Context & 0.470 & 0.626 & 0.682 & 0.636\\ \cline{1-1} \cline{3-7} 
4 &  & Local Context + Speaker Context & \textbf{0.539} & \textbf{0.674} & \textbf{0.727} & \textbf{0.689}\\ \hline \hline
5 & \multirow{4}{*}{BERT Baseline} & - & 0.483 & 0.620 & 0.669 & 0.632\\ \cline{1-1} \cline{3-7}
6 &  & Local Context & \textbf{0.657} & \textbf{0.759} & 0.787 & 0.769\\ \cline{1-1} \cline{3-7}
7 &  & Speaker Context & 0.625 & 0.733 & 0.794 & 0.751\\ \cline{1-1} \cline{3-7} 
8 &  & Local Context + Speaker Context & 0.653 & \textbf{0.759} & \textbf{0.810} & \textbf{0.774}\\ \hline \hline
\end{tabular}
\caption{Results for different experimental settings.  Each row shows the best results for the corresponding settings when varying context size and position. Bold font shows the best results for each model.}
\label{tab:results}
\end{table*}

\begin{figure*}[!t]
\setlength{\abovecaptionskip}{0pt plus 0pt minus 2pt}
\setlength{\belowcaptionskip}{-5pt plus 0pt minus 2pt}
\begin{center}
  \includegraphics[scale=0.33]{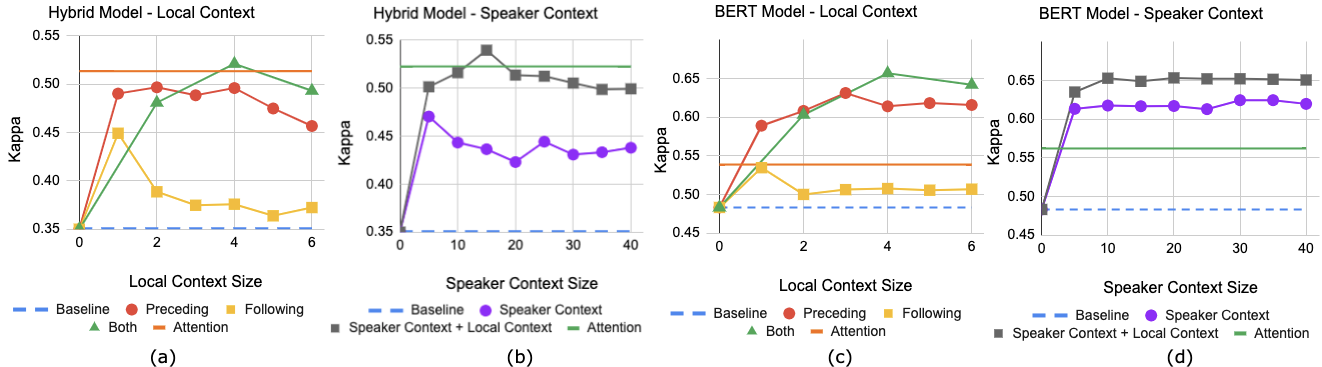}
  \end{center}
  \vspace{-1mm}
  \caption{Results for various context sizes. (a) and (c) compare the baselines to models with local context. (b) and (d) compare the baselines with speaker context and both context types.} 
\label{fig:results}
\vspace{-2mm}
\end{figure*}

\textbf{Local Context.} \quad
We extended both baseline models with local context extracted in three different ways: only ADUs preceding the target, only following ADUs, and both preceding/following ADUs.
We report  three main observations when adding context to the hybrid baseline.
First, with respect to position, all models including prior ADUs in local context significantly outperformed the baseline (p-value $<$ 0.01), while the same is not true for models including next ADUs.
Including both prior and next ADUs, though, resulted in the best performance for local context models (row 2 in Table \ref{tab:results}).
When including only next ADU context,  only context size 1 gave a significant performance improvement over the baseline, while larger context sizes yielded non statistically significant differences.
Second, with respect to context size, the plots in Figure \ref{fig:results} (a) show that
although there is a diminishing return effect as context size increases, increasing context size from 2 (1 prior/next ADU) to 4 (2 prior/next ADUs) results in significantly better precision and f-score (p-value $<$ 0.05).
Overall, we argue that context position and size should both be optimized.
Figure \ref{fig:results} (a) and (c) show that choosing the optimal context position may even be more important than choosing the optimal size: the differences between the three lines are bigger than differences within each line.
Also note that since model complexity increases linearly with context size, larger datasets may be able to take advantage of larger contexts.
Third, results obtained using the attention mechanism are not statistically significantly different than the best results for both local context and speaker context. This indicates that context size and position can be automatically optimized during training time with a marginal performance penalty.
The results obtained from the BERT models confirm our first two observations, as we can see from the similarities in Figures \ref{fig:results} (a) and (c).
For the BERT models, however, the attention mechanism performs significantly worse than any setting that includes at least 1 preceding ADU. In this case context size and position still need to be tuned hyperparameters.
We should also note that overall BERT models perform better than the hybrid models.

\textbf{Speaker Context.} \quad
We define speaker context size K as the K closest prior ADUs to the target ADU.
We experimented with speaker context sizes ranging from a minimum of 5 to a maximum of 40 (effectively considering all speaker turns since the beginning of the discussion).
As shown in Figure \ref{fig:results} (b), including any previous ADUs from the current speaker improves performance over the baseline model.
For hybrid models all context sizes result in statistically significantly better precision, recall, and f-score (p-value $<$ 0.05) compared to the baseline.
We observed a diminishing return effect as speaker context size increases as well, which suggests that a speaker's recent ADUs are perhaps more important than earlier ones.
By comparing the green and red lines in Figure \ref{fig:results} (a) to the purple line in Figure \ref{fig:results} (b) we observed that, taken individually, local context improves performance more than speaker context.
Results also show that the attention mechanism can successfully optimize speaker context size for the hybrid model.
Figure \ref{fig:results} (d) shows that our observations for BERT models are largely consistent with those for hybrid models, though we note that BERT models seem less sensitive to speaker context size.
Unfortunately, adding attention mechanisms to optimize speaker context size results in a large performance penalty, though still outperforming the baseline model.

\textbf{Local Context and Speaker Context.} \quad
After individually adding each of the two context types to the baseline models, we experimented with including both context types simultaneously.
In this setting we obtained the best overall performance (rows 4 and 8 of Table \ref{tab:results}).
As we can see from the grey and purple lines in Figure \ref{fig:results} (b), for hybrid models, modeling both contexts simultaneously always outperforms speaker context alone. For speaker context size $>$ 5, improvements are statistically significant.
In this experiment we kept local context size and position constant (both prior and next ADUs, size 4) and varied speaker context size.
Repeating this experiment with different local context settings yielded similar results.
We also observed that by combining both context types we were able to achieve kappa $>$ 0.5 and f1-score $>$ 0.65 more consistently than with either context type individually.
We observed consistent trends for BERT models, although the differences between speaker context alone and local context + speaker context are often not statistically significant.

\section{Conclusions and Future Work}
\label{sec:conclusions}
In this paper we analyzed the impact of context for predicting argument components in multi-party discussions.
We defined two types of context, local context and speaker context, and analyzed how different models perform when varying context size. We also investigated the effect of different positions for local context.
We performed evaluations of the two context types separately as well as simultaneously on two types of neural network models.
Experimental results support our claim that both context size and position are important when incorporating context in argument mining systems, therefore sentences beyond the ones immediately surrounding an ADU should be considered.
Our results also show that speaker context can improve performance in multi-party discussions.
Finally, we investigated the use of an attention mechanism for optimizing context size and found that its effectiveness is dependent on the type of model used.
Our future plans include two main directions: 1)  repeating our local context experiments on larger datasets, including other domains; 2) evaluating the effectiveness of speaker context on multi-party web discussions, where discussions are usually longer and author ID is typically available.

\section*{Acknowledgements}

This work was supported by the National Science Foundation (1842334 and 1917673),
and in part by the University of Pittsburgh Center for Research Computing
through the resources provided.



\bibliographystyle{coling}
\bibliography{coling2020}

\begin{thebibliography}{}

\bibitem[\protect\citename{Aker \bgroup et al.\egroup }2017]{Aker:17}
Ahmet Aker, Alfred Sliwa, Yuan Ma, Ruishen Lui, Niravkumar Borad, Seyedeh
  Ziyaei, and Mina Ghobadi.
\newblock 2017.
\newblock What works and what does not: Classifier and feature analysis for
  argument mining.
\newblock In {\em Proceedings of the 4th Workshop on Argument Mining}, pages
  91--96.

\bibitem[\protect\citename{Chakrabarty \bgroup et al.\egroup
  }2019]{Chakrabarty:19}
Tuhin Chakrabarty, Christopher Hidey, Smaranda Muresan, Kathy McKeown, and
  Alyssa Hwang.
\newblock 2019.
\newblock {AMPERSAND}: Argument mining for {PERS}u{A}sive o{N}line discussions.
\newblock In {\em Proceedings of the 2019 Conference on Empirical Methods in
  Natural Language Processing and the 9th International Joint Conference on
  Natural Language Processing (EMNLP-IJCNLP)}, pages 2933--2943, Hong Kong,
  China, November.

\bibitem[\protect\citename{Devlin \bgroup et al.\egroup }2019]{devlin:18}
Jacob Devlin, Ming-Wei Chang, Kenton Lee, and Kristina Toutanova.
\newblock 2019.
\newblock {BERT}: Pre-training of deep bidirectional transformers for language
  understanding.
\newblock In {\em Proceedings of the 2019 Conference of the North {A}merican
  Chapter of the Association for Computational Linguistics: Human Language
  Technologies}, pages 4171--4186, Minneapolis, Minnesota, June.

\bibitem[\protect\citename{Eger \bgroup et al.\egroup }2017]{Eger:17}
Steffen Eger, Johannes Daxenberger, and Iryna Gurevych.
\newblock 2017.
\newblock Neural end-to-end learning for computational argumentation mining.
\newblock In {\em Proceedings of the 55th Annual Meeting of the Association for
  Computational Linguistics (Volume 1: Long Papers)}, pages 11--22.

\bibitem[\protect\citename{Habernal and Gurevych}2017]{Habernal:17}
Ivan Habernal and Iryna Gurevych.
\newblock 2017.
\newblock Argumentation mining in user-generated web discourse.
\newblock {\em Computational Linguistics}, 43(1):125--179.

\bibitem[\protect\citename{Hochreiter and Schmidhuber}1997]{LSTM:97}
Sepp Hochreiter and J{\"u}rgen Schmidhuber.
\newblock 1997.
\newblock Long short-term memory.
\newblock {\em Neural computation}, 9(8):1735--1780.

\bibitem[\protect\citename{Li and Nenkova}2015]{Li:15}
Junyi~Jessy Li and Ani Nenkova.
\newblock 2015.
\newblock Fast and accurate prediction of sentence specificity.
\newblock In {\em Proceedings of the Twenty-Ninth Conference on Artificial
  Intelligence (AAAI)}, pages 2281--2287, January.

\bibitem[\protect\citename{Lugini and Litman}2018]{Lugini:18b}
Luca Lugini and Diane Litman.
\newblock 2018.
\newblock Argument component classification for classroom discussions.
\newblock In {\em Proceedings of the 5th Workshop on Argument Mining}, pages
  57--67.

\bibitem[\protect\citename{Luong \bgroup et al.\egroup }2015]{luong:15}
Minh-Thang Luong, Hieu Pham, and Christopher~D Manning.
\newblock 2015.
\newblock Effective approaches to attention-based neural machine translation.
\newblock In {\em Proceedings of the 2015 Conference on Empirical Methods in
  Natural Language Processing}, pages 1412--1421.

\bibitem[\protect\citename{Nguyen and Litman}2016]{Nguyen:16}
Huy Nguyen and Diane~J Litman.
\newblock 2016.
\newblock Improving argument mining in student essays by learning and
  exploiting argument indicators versus essay topics.
\newblock In {\em FLAIRS Conference}, pages 485--490.

\bibitem[\protect\citename{Olshefski \bgroup et al.\egroup }2020]{olshefski:20}
Christopher Olshefski, Luca Lugini, Ravneet Singh, Diane Litman, and Amanda
  Godley.
\newblock 2020.
\newblock The discussion tracker corpus of collaborative argumentation.
\newblock In {\em Proceedings of The 12th Language Resources and Evaluation
  Conference}, pages 1033--1043, Marseille, France, May. European Language
  Resources Association.

\bibitem[\protect\citename{Opitz and Frank}2019]{Optiz:19}
Juri Opitz and Anette Frank.
\newblock 2019.
\newblock Dissecting content and context in argumentative relation analysis.
\newblock In {\em Proceedings of the 6th Workshop on Argument Mining}, pages
  25--34, Florence, Italy, August.

\bibitem[\protect\citename{Persing and Ng}2016]{Persing:16}
Isaac Persing and Vincent Ng.
\newblock 2016.
\newblock End-to-end argumentation mining in student essays.
\newblock In {\em Proceedings of the 2016 Conference of the North American
  Chapter of the Association for Computational Linguistics: Human Language
  Technologies}, pages 1384--1394.

\bibitem[\protect\citename{Stab and Gurevych}2014]{Stab:14b}
Christian Stab and Iryna Gurevych.
\newblock 2014.
\newblock Annotating argument components and relations in persuasive essays.
\newblock In {\em Proceedings of COLING 2014, the 25th International Conference
  on Computational Linguistics: Technical Papers}, pages 1501--1510.

\bibitem[\protect\citename{Stone and Hunt}1963]{Stone:63}
Philip~J Stone and Earl~B Hunt.
\newblock 1963.
\newblock A computer approach to content analysis: studies using the general
  inquirer system.
\newblock In {\em Proceedings of the May 21-23, 1963, spring joint computer
  conference}, pages 241--256. ACM.

\bibitem[\protect\citename{Toulmin}1958]{Toulmin:58}
Stephen Toulmin.
\newblock 1958.
\newblock {\em The uses of argument}.
\newblock Cambridge: Cambridge University Press.

\bibitem[\protect\citename{Vaswani \bgroup et al.\egroup }2017]{vaswani:17}
Ashish Vaswani, Noam Shazeer, Niki Parmar, Jakob Uszkoreit, Llion Jones,
  Aidan~N Gomez, {\L}ukasz Kaiser, and Illia Polosukhin.
\newblock 2017.
\newblock Attention is all you need.
\newblock In {\em Advances in neural information processing systems}, pages
  5998--6008.

\bibitem[\protect\citename{Wilson \bgroup et al.\egroup }2009]{Wilson:09}
Theresa Wilson, Janyce Wiebe, and Paul Hoffmann.
\newblock 2009.
\newblock Recognizing contextual polarity: An exploration of features for
  phrase-level sentiment analysis.
\newblock {\em Computational linguistics}, 35(3):399--433.

\bibitem[\protect\citename{Wilson}1988]{Wilson:88}
Michael Wilson.
\newblock 1988.
\newblock Mrc psycholinguistic database: Machine-usable dictionary, version
  2.00.
\newblock {\em Behavior Research Methods}, 20(1):6--10.

\bibitem[\protect\citename{Wolf \bgroup et al.\egroup
  }2019]{HuggingFacesTransformers}
Thomas Wolf, Lysandre Debut, Victor Sanh, Julien Chaumond, Clement Delangue,
  Anthony Moi, Pierric Cistac, Tim Rault, R'emi Louf, Morgan Funtowicz, and
  Jamie Brew.
\newblock 2019.
\newblock Huggingface's transformers: State-of-the-art natural language
  processing.
\newblock {\em ArXiv}, abs/1910.03771.

\end{thebibliography}

\end{document}